%% file: main.tex
\documentclass[10pt,twocolumn,letterpaper]{article}

\usepackage{cvpr}              
\usepackage{placeins}
\input{preamble}

\definecolor{cvprblue}{rgb}{0.21,0.49,0.74}
\usepackage[pagebackref,breaklinks,colorlinks,allcolors=cvprblue]{hyperref}

\def\paperID{} 
\def\confName{CVPR}
\def\confYear{2026}

\title{Tutor-Student Reinforcement Learning: A Dynamic Curriculum for Robust Deepfake Detection}

\author{Zhanhe Lei$^1$, Zhongyuan Wang$^{1}$\thanks{Corresponding author}, Jikang Cheng$^2$, Baojin Huang$^3$, \\
    Yuhong Yang$^1$, Zhen Han$^1$$^{ *}$, Chao Liang$^1$, Dengpan Ye$^4$ \\
    School of Computer Science, Wuhan University$^1$ \\
     School of Integrated Circuits, Peking University$^2$\\
    School of Information, Huazhong Agricultural University$^3$\\
    Cyberspace Institute of Advanced Technology, Guangzhou University$^4$\\
    {\tt\small zhanhelei@whu.edu.cn}}

\begin{document}
\maketitle
\input{sec/0_abstract}    
\input{sec/1_intro}
\input{sec/2_relatedwork}

\input{sec/3_methodology}
\input{sec/4_Experiment}
\input{sec/5_Conclusions}
\input{sec/6_Acknowledgments_and_Disclosure_of_Funding}

\FloatBarrier
{
    \small
    \bibliographystyle{ieeenat_fullname}
    \bibliography{main}
}
\end{document}

%% file: preamble.tex
\usepackage[dvipsnames]{xcolor} 

\usepackage{amsmath}   
\usepackage{amssymb}   
\usepackage{multirow}
\usepackage[utf8]{inputenc}
\usepackage{booktabs} 
\usepackage{dcolumn}  
\usepackage{lipsum}   
\usepackage{tabularx}
\usepackage{subcaption}
\usepackage{microtype}
\usepackage{algorithm}
\usepackage{algpseudocode}

\usepackage{amsmath}
\usepackage{amssymb}

\newcolumntype{d}[1]{D{.}{.}{#1}}

%% file: sec/0_abstract.tex
\begin{abstract}
Standard supervised training for deepfake detection treats all samples with uniform importance, which can be suboptimal for learning robust and generalizable features. In this work, we propose a novel Tutor-Student Reinforcement Learning (TSRL) framework to dynamically optimize the training curriculum. Our method models the training process as a Markov Decision Process where a ``Tutor'' agent learns to guide a ``Student'' (the deepfake detector). The Tutor, implemented as a Proximal Policy Optimization (PPO) agent, observes a rich state representation for each training sample, encapsulating not only its visual features but also its historical learning dynamics, such as EMA loss and forgetting counts. Based on this state, the Tutor takes an action by assigning a continuous weight (0-1) to the sample's loss, thereby dynamically re-weighting the training batch. The Tutor is rewarded based on the Student's immediate performance change, specifically rewarding transitions from incorrect to correct predictions. This strategy encourages the Tutor to learn a curriculum that prioritizes high-value samples, such as hard-but-learnable examples, leading to a more efficient and effective training process. We demonstrate that this adaptive curriculum improves the Student's generalization capabilities against unseen manipulation techniques compared to traditional training methods. Code is available at \textit{\href{https://github.com/wannac1/TSRL}{https://github.com/wannac1/TSRL}}.
\end{abstract}

%% file: sec/1_intro.tex
\section{Introduction}
\label{sec:intro}
\begin{figure}[htbp]
  \centering
  \includegraphics[width=\linewidth]{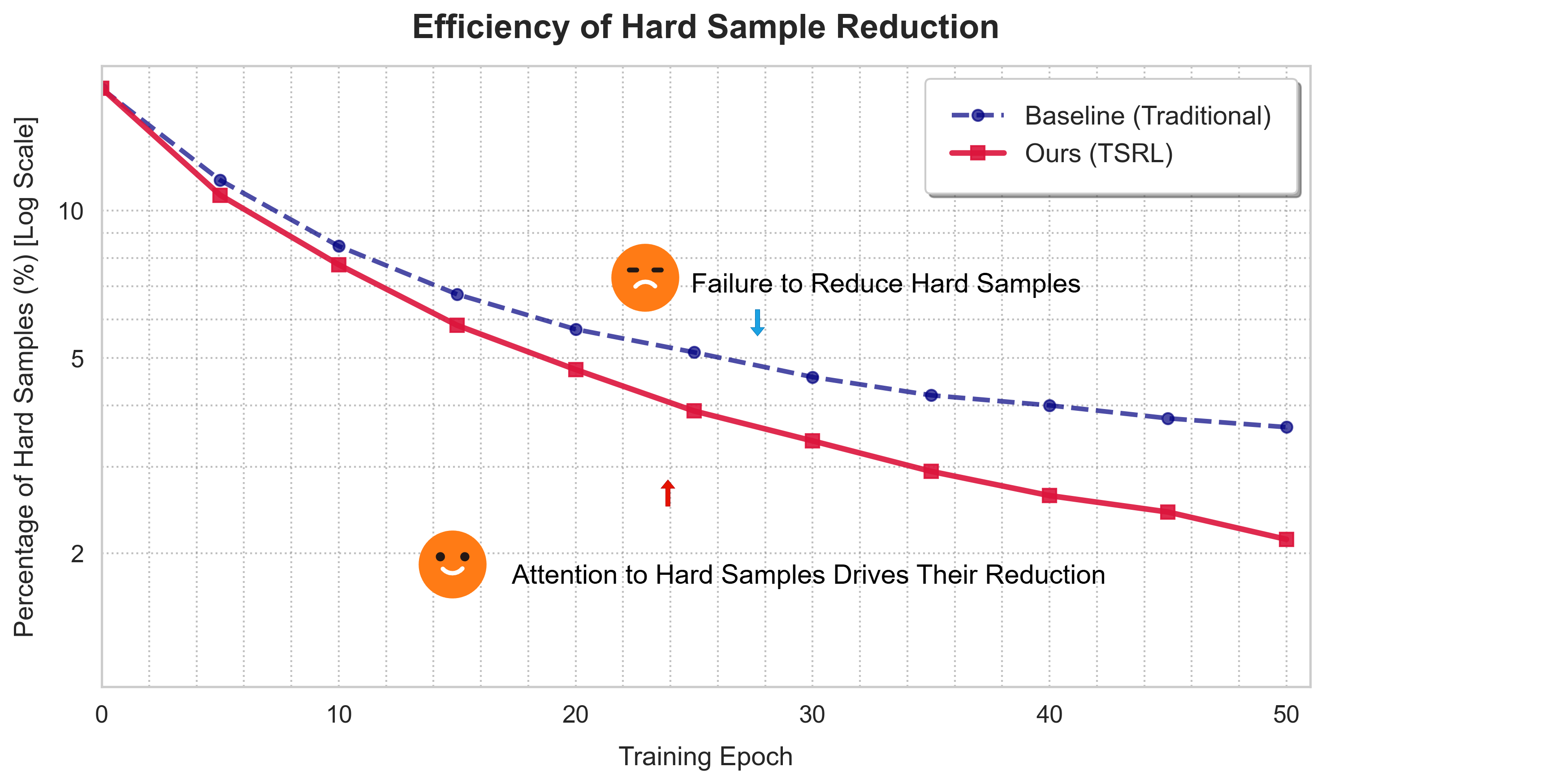}
  \caption{Comparison of hard sample (Historical EMA Loss $>$ 0.7) reduction efficiency during training. Our TSRL framework (red) demonstrates a significantly faster convergence in resolving difficult samples compared to the traditional baseline (blue), as highlighted by the logarithmic scale.}
  \label{fig:hard_sample_comparison}
\end{figure}

The proliferation of high-fidelity, AI-generated media, or ``deepfakes'', presents a significant threat to information integrity, personal security, and public trust \cite{wang2023survey, masood2023deepfakes}. In response, the development of robust deepfake detection methods has become a critical area of research \cite{radford2021learning, cheng2024can, yan2023ucf, yan2024transcending, lin2024fake, xu2023tall, huang2023implicit, shiohara2022detecting, cao2022end, Cheng_2025_CVPR, chen2022self, ni2022core, luo2021generalizing, liu2021spatial, Cheng_tip}. While current state-of-the-art detectors can achieve high accuracy on known datasets, their performance often degrades significantly when faced with novel manipulation techniques, unseen compression artifacts, or different data domains \cite{wang2023survey}. This poor generalization remains the foremost challenge in the field.

We attribute this failing, in part, to the conventional supervised training paradigm. The approach of applying a uniform loss gradient across all samples is suboptimal. Recent work has highlighted that not all samples contribute equally to training; for instance, the quality of AI-generated images can significantly impact a detector's performance, suggesting that samples should be weighted differently \cite{song2024quality, xiao2024highquality}. Indeed, this conventional approach is inefficient, struggling to reduce the number of difficult samples over time as depicted by the persistently high percentage of hard samples in the Baseline (Traditional) curve shown in Figure~\ref{fig:hard_sample_comparison}. This inability of baseline models to effectively resolve hard samples often correlates with compromised generalization capabilities. However, these approaches primarily investigate \textit{static} weightings based on predetermined sample properties like image quality. We argue that an effective training curriculum should be \textit{dynamic}, adapting not only to the samples themselves but also to the evolving state of the model as it learns. This concept draws inspiration from Curriculum Learning \cite{bengio2009curriculum, lin2024fake}, which originally posited that models learn better from examples presented in a meaningful order, such as by increasing difficulty. Our approach extends this idea; instead of pre-sorting the data, our framework learns a dynamic policy to \textit{re-weight} samples within each batch, effectively creating a real-time curriculum based on the model's instantaneous learning state. In stark contrast, our TSRL approach, by dynamically allocating more attention to these challenging samples, effectively accelerates their reduction throughout the training epochs, as evidenced by the significantly lower and continuously decreasing red curve in Figure~\ref{fig:hard_sample_comparison}. This dynamic prioritization ultimately leads to a more robust and generalized model.

\begin{figure*}[t]
  \centering
  \includegraphics[width=\textwidth]{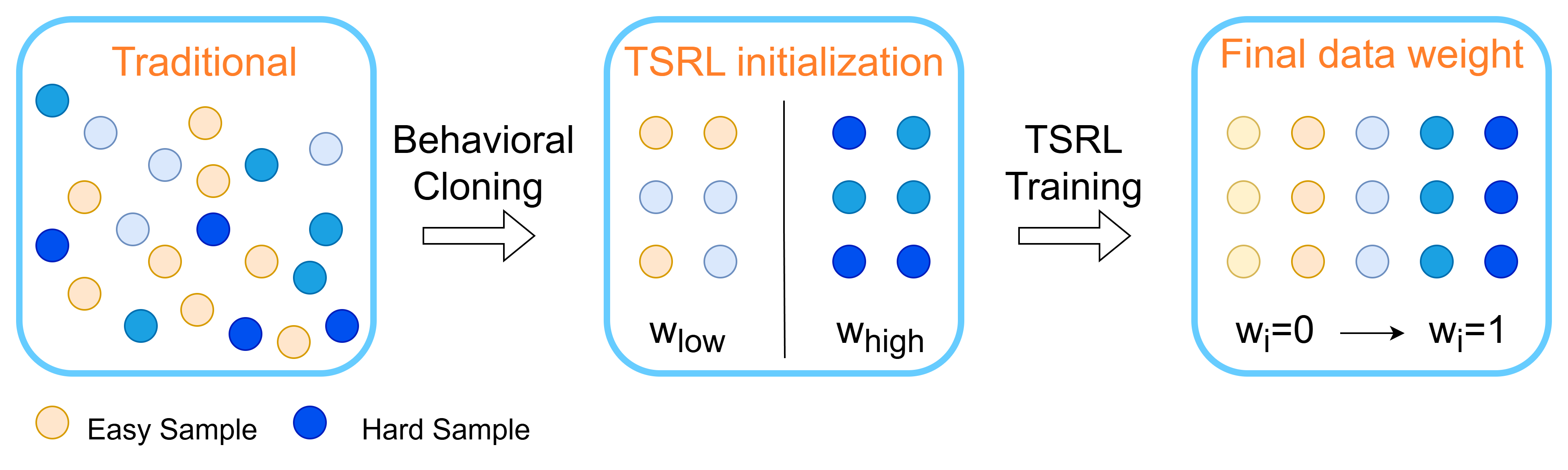}
  \caption{A simplified overview of our proposed Tutor-Student Reinforcement Learning (TSRL) framework. The Tutor (RL Agent) learns a policy to dynamically assign weights to training samples (e.g., up-weighting ``Hard'' samples, down-weighting ``Easy'' samples) to optimize the Student (Detector) for generalization.}
  \label{fig:tsrl_intro}
\end{figure*}

To overcome the limitations of static training, we re-frame the process not as a one-size-fits-all optimization problem, but as a dynamic teaching problem. We propose a novel Tutor-Student Reinforcement Learning (TSRL) framework, where a ``Student'' (the deepfake detector) is guided by an intelligent ``Tutor'' (an RL agent). This Tutor learns an optimal \textit{teaching policy}, or curriculum, to maximize the Student's learning and generalization (see Figure~\ref{fig:tsrl_intro}). Our method models the training procedure as a Markov Decision Process (MDP). At each training step, the Tutor agent observes a comprehensive state for each sample in the batch. This state is a rich vector encapsulating not only the Student's current visual feature representation of the sample but also its unique learning history, such as its exponential moving average (EMA) loss and historical ``forgetting counts''. Based on this state, the Tutor, implemented as a Proximal Policy Optimization (PPO) agent \cite{schulman2017ppo}, takes an action: it assigns a continuous weight between 0 and 1 to that sample's loss. This action allows the Tutor to dynamically re-shape the loss landscape, forcing the Student to ``pay attention'' to samples the Tutor deems most valuable for learning at that exact moment. To learn this policy, the Tutor receives a reward based on the Student's \textit{immediate performance change}---a dense signal that provides high rewards for actions that cause the Student to flip an incorrect prediction to a correct one. This encourages the Tutor to discover a sophisticated curriculum that might, for example, prioritize hard-but-learnable examples while ignoring samples that are either too easy or hopelessly confusing.

In this paper, we demonstrate that TSRL framework, by creating an adaptive and dynamic curriculum, trains a deepfake detector that is more efficient and, most importantly, exhibits significantly improved generalization against unseen manipulation techniques. Our contributions are as follows:
\begin{enumerate}[label=\textbullet, leftmargin=*, itemsep=2pt, topsep=3pt]
    \item \textbf{TSRL Framework.} We propose a novel Tutor-Student framework for deepfake detection that models the training process as an MDP.
    \item \textbf{History-Aware State.} We design a rich, history-aware state representation that allows an agent to make informed curriculum decisions.
    \item \textbf{State-Change Reward.} We introduce a state-change-based reward function that effectively captures immediate learning progress to guide the Tutor.
    \item \textbf{Superior Generalization.} We provide empirical validation demonstrating that our framework significantly improves detector generalization in both Cross-dataset and Cross-method evaluations.
\end{enumerate}

%% file: sec/2_relatedwork.tex
\section{Related Work}
\label{sec:related_work}

Our research is positioned at the intersection of deepfake detection and reinforcement learning. We first review relevant advances in deepfake detection, particularly concerning generalization, before discussing the reinforcement learning principles that form the basis of our novel training paradigm.

\subsection{Deepfake Detection}
The field of deepfake detection has rapidly evolved to counter the increasing sophistication of generative models \cite{goodfellow2014generative}. Comprehensive surveys \cite{wang2023survey, masood2023deepfakes} detail this co-evolution, from early methods targeting specific visual artifacts \cite{afchar2018mesonet} to modern approaches designed to capture more subtle inconsistencies.

A primary challenge in deepfake detection is generalization to unseen forgery methods. To find more robust forgery traces, researchers have explored various signal domains and learning strategies. One line of inquiry focuses on specific artifacts, such as analyzing frequency-domain inconsistencies \cite{luo2021generalizing, liu2021spatial, li2021frequency}, using reconstruction-based methods to identify anomalous regions \cite{cao2022end, ni2022core, zhu2021face}, or focusing on blending boundaries, often through self-blended image training \cite{shiohara2022detecting, chen2022self, li2020facexray, zhao2021learning}. Other methods exploit spatio-temporal inconsistencies \cite{xu2023tall} or disrupted biometric signals \cite{huang2023implicit}. More recent work has shifted focus to learning explicitly generalizable representations. This includes methods to uncover common, forgery-agnostic features \cite{yan2023ucf}, use latent space augmentations to simulate a wider variety of fakes \cite{yan2024transcending}, and train with self-supervised adversarial examples \cite{chen2022self}. Some approaches even attempt to train without any deepfake data at all \cite{cheng2024can}, often by leveraging large-scale pre-training \cite{radford2021learning} to learn robust priors.

Several studies note that the quality of fakes impacts detector performance \cite{song2024quality, xiao2024highquality}, suggesting that the model should pay more attention to high-quality, difficult-to-classify instances. This has led to the proposal of curriculum-based training, for instance, by progressively increasing the difficulty of forgery augmentations \cite{lin2024fake}. Our work builds directly on this observation; however, rather than using a static or pre-defined curriculum, we employ a dynamic, adaptive policy learned through reinforcement learning.

\subsection{Reinforcement Learning and Curriculum Learning}
The concept of Curriculum Learning (CL) \cite{bengio2009curriculum}, which posits that models learn more effectively from data presented in a meaningful order (e.g., easy-to-hard), provides a strong foundation for our work. We frame the task of ordering or weighting training data as a policy optimization problem, which can be solved using Reinforcement Learning (RL). In this paradigm, an RL agent learns a ``policy'' to maximize a cumulative ``reward''. Policy gradient methods, such as Proximal Policy Optimization (PPO) \cite{schulman2017ppo}, are particularly effective for training agents in complex environments.

While the application of RL to deepfake detection is nascent, its potential has been explored in adjacent tasks. For instance, RL has been used to learn optimal data augmentation policies to improve cross-dataset generalization \cite{nadimpalli2022improving}. More recently, RL-based strategies have been employed to optimize large multimodal models for related tasks, such as generating explanations for detections \cite{li2025raidx} or benchmarking detection capabilities \cite{wang2025dfbench}. However, to our knowledge, no prior work has utilized RL to learn a dynamic, sample-weighting policy that adapts the training curriculum for a deepfake detector in real-time. Our work presents this novel formulation: we are the first to design a ``Tutor'' agent that uses RL to learn a dynamic, sample-weighting policy for a ``Student'' deepfake detector. This Tutor observes the Student's learning state and adjusts the curriculum in real-time to maximize generalization, moving beyond static curricula or more general RL applications.

%% file: sec/3_methodology.tex
\begin{figure*}[t]
    \centering
    \includegraphics[width=\textwidth]{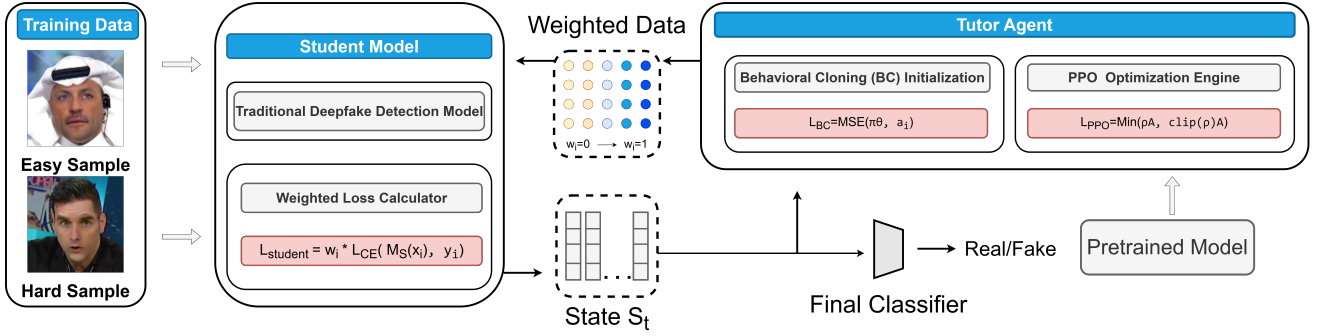}
    \caption{The Tutor-Student Reinforcement Learning (TSRL) Framework}
    \label{fig:tsrl_framework}
\end{figure*}
\section{Methodology}
\label{sec:Methodology}
\subsection{Motivations}
Deepfake detection models are predominantly trained using standard supervised learning, which typically relies on empirical risk minimization. This traditional paradigm assigns uniform importance to all training samples, applying an equal loss weight irrespective of a sample's individual utility to the learning process.
While this approach yields high fidelity on in-distribution test data, it concurrently results in a significant generalization deficit when models are evaluated on unseen forgery methods, compression artifacts, or other perturbations. This generalization gap remains a critical vulnerability. Some methods attempt to mitigate this by employing Curriculum Learning (CL) \cite{lin2024fake}, which organizes training data based on a pre-defined difficulty metric. And another dynamic curriculum learning method \cite{song2024towards} relies on rule-based, monotonic pacing (e.g., sine wave), blindly following a schedule regardless of model status. 

However, such static curricula are inherently limited. The difficulty of a given sample is not an intrinsic, constant property but is relative to the detector's instantaneous learning state. A sample that is complex for a nascent model may become trivial as training progresses, while other samples may remain persistently challenging. A static, model-agnostic curriculum cannot adapt to this dynamic learning process. It may inefficiently expend computational resources on ``easy'' samples that the model has already mastered, leading to a neglect of ``hard'' samples that are crucial for refining the discriminative boundary. This imbalance can bias the model towards superficial, overfitted features rather than robust, generalizable forgery traces.

We hypothesize that a dynamic training strategy, which modulates the training curriculum in real-time based on the detector's evolving state, will cultivate superior generalization. We propose to formalize this adaptive curriculum generation as a sequential decision-making process. This allows us to leverage Reinforcement Learning (RL) to train a ``Tutor'' agent. This agent's objective is to learn an optimal, dynamic sample-weighting policy for a ``Student'' detector, where the explicit optimization goal is the maximization of the Student's generalization performance on out-of-distribution validation data.

\subsection{Tutor-Student RL (TSRL) Framework}
To \textbf{operationalize} the dynamic, adaptive curriculum outlined in our motivation, we conceptualize and formalize the training procedure as a Markov Decision Process (MDP). This formalization is ideally suited for modeling the \textbf{sequential nature} of selecting optimal training data based on the model's current state. To solve this MDP, we introduce the Tutor-Student Reinforcement Learning (TSRL) framework. As illustrated in Figure~\ref{fig:tsrl_framework}, TSRL employs a `Tutor' (an RL agent) that learns an adaptive policy to guide the training of the `Student' (our main model), thereby \textbf{dynamically tailoring} the curriculum to the Student's evolving capabilities.
The framework consists of two primary agents: the \textbf{Student} ($M_S$), which is the Deepfake detector being trained, and the \textbf{Tutor} ($T_\pi$), which is an RL agent tasked with learning the optimal training policy $\pi$. A key third component is the \textbf{State Manager} ($SM$), which maintains the longitudinal learning history for every sample in the training set.
The TSRL interaction loop, executed at each training step $t$ for a given sample $x_i$, is defined by its State, Action, and Reward.

\paragraph{State ($s_t$).}
The state $s_t$ is a comprehensive vector designed to provide the Tutor with a complete snapshot of both the sample's current difficulty and its historical learning trajectory. This state $s_i$ for a sample $x_i$ is a concatenation of current and historical features:
\begin{equation}
s_i = [f_i, p_i, e_i, l_i^{ema}, c_i^{forget}],
\end{equation}
where the components are defined as:
\begin{itemize}
\item $f_i$: The deep feature vector extracted from an intermediate layer of the Student model $M_S$.
\item $p_i$: The Student's current prediction confidence for the target class.
\item $e_i$: A one-hot vector indicating if the Student's current prediction for $x_i$ is correct.
\item $l_i^{ema}$: The normalized Exponential Moving Average (EMA) of the sample's loss over past epochs. This represents the sample's long-term perceived difficulty. The EMA loss $l_i^{ema}$ at epoch $t$ is calculated recursively:
\begin{equation}
l_i^{ema}(t) = \beta \cdot l_i^{ema}(t-1) + (1 - \beta) \cdot \mathcal{L}_{\text{CE}}(M_S(x_i), y_i), 
\end{equation}
where $\beta$ is the smoothing factor.
\item $c_i^{forget}$: The normalized count of ``forgetting events'' for sample $x_i$. This metric tracks how many times the model correctly classified $x_i$ after having failed on it in the previous epoch, explicitly measuring learning instability.
\end{itemize}
This state definition is critical, as it allows the Tutor to differentiate between a sample that is consistently difficult (high $l_i^{ema}$) and one that is unstably learned (high $c_i^{forget}$).

\paragraph{Action ($a_t$).}
Given the state $s_i$, the Tutor agent $T_\pi$ samples an action $a_i$ from its policy $\pi(a_i | s_i)$. The action $a_i$ is a continuous scalar value $w_i \in [0, 1]$, achieved via a $\mathrm{Sigmoid}$ activation $\sigma(\cdot)$ on the policy network's raw output logit $z_i$:
\begin{equation}
w_i = \sigma(z_i) = \frac{1}{1 + e^{-z_i}}. 
\end{equation}
This action $w_i$ is applied as a weight to the standard supervised loss for that sample. The Student's training objective for sample $(x_i, y_i)$ is thus a weighted cross-entropy loss:
\begin{equation}
\mathcal{L}_{\text{student}} = w_i \cdot \mathcal{L}_{\text{CE}}(M_S(x_i), y_i). 
\end{equation}
This operation allows the Tutor to adaptively adjust the learning focus, dynamically up-weighting'' challenging samples and down-weighting'' trivial ones. This process steers the Student's gradient updates to be driven by the most informative data, maximizing the impact of each training step.

\paragraph{Reward ($r_t$).}
To optimize its policy, the Tutor requires a reward signal that measures the utility of its last action $w_i$. A naive reward based on final validation accuracy would be sparse and delayed. Instead, we formulate a dense, immediate reward based on the \emph{instantaneous change} in the Student's performance, measured directly before and after the weighted gradient update $M_S \rightarrow M_S'$.
This reward is defined as:
\begin{equation} \label{eq:reward}
r_i = 
\begin{cases} 
  +1.0 & \text{Error $\to$ Correct} \quad (\neg c_{\mathrm{init}} \land c_{\mathrm{upd}}) \\
  -1.0 & \text{Correct $\to$ Error} \quad (c_{\mathrm{init}} \land \neg c_{\mathrm{upd}}) \\
  c_{\mathrm{rew}} \cdot \Delta_{\mathrm{conf}} & \text{Correct $\to$ Correct} \quad (c_{\mathrm{init}} \land c_{\mathrm{upd}}) \\
  -c_{\mathrm{rew}} \cdot \Delta_{\mathrm{conf}} & \text{Error $\to$ Error} \quad (\neg c_{\mathrm{init}} \land \neg c_{\mathrm{upd}}), 
\end{cases}
\end{equation}
where $\Delta_{\mathrm{conf}} = \mathrm{conf}_{\mathrm{upd}} - \mathrm{conf}_{\mathrm{init}}$, $c_{\text{init}}$ and $\text{conf}_{\text{init}}$ are the correctness and confidence of $M_S$ before the update, $c_{\text{upd}}$ and $\text{conf}_{\text{upd}}$ are the values after the update $M_S'$, and $c_{\text{rew}}$ is a scaling coefficient. This ``state-change'' reward directly incentivizes the Tutor to assign weights that cause immediate learning progress: flipping incorrect predictions to correct, stabilizing correct ones at higher confidence, and penalizing actions that cause forgetting or increase confidence in an incorrect prediction.

\subsection{Tutor Training}
\label{sec:TSRL}

The complete training process is structured in three phases to ensure stability and effective policy convergence:
\begin{enumerate}[label=\textbullet, leftmargin=*, itemsep=2pt, topsep=3pt]
\item \textbf{Behavioral Cloning (BC) Initialization}: Training an RL agent from a random policy is notoriously unstable and sample-inefficient. We first pre-train the Tutor agent using Behavioral Cloning to provide a stable starting policy. An ``expert'' policy is defined heuristically, designed to mimic a simple self-paced learning strategy (e.g., favoring samples with moderate loss). A dataset $\mathcal{D}$ of $(s_i, a_i^{\text{expert}})$ pairs is generated from this heuristic, and the Tutor's policy network $\pi_\theta$ is pre-trained via supervised learning to mimic this behavior. The BC loss function $\mathcal{L}_{BC}$ is a Mean Squared Error:
\begin{equation}
\mathcal{L}_{BC}(\theta) = \mathbb{E}_{(s_i, a_i^{\text{expert}}) \sim \mathcal{D}} \left[ (\pi_\theta(s_i) - a_i^{\text{expert}})^2 \right].
\end{equation}
This initialization provides the Tutor with a non-random baseline policy before commencing online RL training.

\item \textbf{Student Warmup}: The main training loop begins by first training the Student $M_S$ with standard supervised learning (all $w_i = 1.0$) for $N_{\text{warmup}}$ epochs. This warmup phase is critical, as it allows the Student detector to move beyond its initial random parameter space. This, in turn, enables the state manager to accumulate meaningful initial statistics, ensuring the state vector $s_t$ is reliable and stable before the Tutor begins active decision-making.

\item \textbf{TSRL Training}: After the warmup, the full MDP loop (State $\to$ Action $\to$ Weighted Update $\to$ Reward) is executed for each batch. All experiences $(s_t, a_t, r_t, \log \pi(a_t|s_t))$ are collected in a rollout buffer. At the end of each epoch, the Tutor's policy $\pi$ is updated using this collected batch of experiences via the Proximal Policy Optimization (PPO) algorithm \cite{schulman2017ppo}. PPO is a policy gradient method known for its stability, optimizing a clipped surrogate objective function, $\mathcal{L}_{\text{PPO}}$, defined as:
\begin{equation}
\mathcal{L}_{\text{PPO}}(\theta) = \mathbb{E}_t \left[ \min \left( \rho_t(\theta) \hat{A}_t, \text{clip}(\rho_t(\theta), 1-\epsilon, 1+\epsilon) \hat{A}_t \right) \right],
\end{equation}
where $\rho_t(\theta) = \frac{\pi_\theta(a_t|s_t)}{\pi_{\theta_{\text{old}}}(a_t|s_t)}$ is the probability ratio, $\hat{A}_t$ is the estimated advantage function (e.g., using GAE), and $\epsilon$ is the clipping hyperparameter.

\end{enumerate}
This combination of BC pre-training for stable initialization, Student warmup for reliable state generation, and PPO training with a dense state-change reward signal allows the Tutor agent to stably and efficiently learn a sophisticated, dynamic curriculum far superior to any static heuristic.

%% file: sec/4_Experiment.tex
\section{Experiments} 

\begin{table*} [t]
\centering
\caption{\textbf{Cross-dataset and Cross-method Evaluation.} All models are trained on FF++ (c23) \cite{rossler2019faceforensics} and evaluated on the other datasets/fake data. Results from models trained with our TSRL are highlighted. $\dagger$ denotes models retrained by us; results for all other methods are sourced from \cite{yan2024effort, cheng2024can, DeepfakeBench_YAN_NEURIPS2023}.}
\label{tab:my_combined_evaluation} 
\footnotesize 
\setlength{\tabcolsep}{1pt} 

\begin{tabular*}{\textwidth}{l@{\extracolsep{\fill}} ccccc | ccccccccc} 
\toprule

\multirow{2} {*} {Methods} & \multicolumn{5} {c|} {Cross-dataset Evaluation} & \multicolumn{9} {c} {Cross-method Evaluation} \\ 
\cmidrule (lr){2-6} \cmidrule (lr){7-15}
& CDF-v2 & DFD & DFDC & DFDCP & Avg. & UniFace & BleFace & MobSwap & e4s & FaceDan & FSGAN & InSwap & SimSwap & Avg. \\
\midrule
F3Net \cite{qian2020thinking} & 0.789 & 0.844 & 0.718 & 0.749 & 0.775 & 0.809 & 0.808 & 0.867 & 0.494 & 0.717 & 0.845 & 0.757 & 0.674 & 0.746 \\
SPSL \cite{liu2021spatial} & 0.799 & 0.871 & 0.724 & 0.770 & 0.791 & 0.747 & 0.748 & 0.885 & 0.514 & 0.666 & 0.812 & 0.643 & 0.665 & 0.710 \\
SRM \cite{luo2021generalizing} & 0.840 & 0.885 & 0.695 & 0.728 & 0.787 & 0.749 & 0.704 & 0.779 & 0.704 & 0.659 & 0.772 & 0.793 & 0.694 & 0.732 \\
RECCE \cite{cao2022end} & 0.823 & 0.891 & 0.696 & 0.734 & 0.786 & 0.898 & 0.832 & 0.925 & 0.683 & 0.848 & 0.949 & 0.848 & 0.768 & 0.844 \\
SLADD \cite{chen2022self} & 0.837 & 0.904 & 0.772 & 0.756 & 0.817 & 0.878 & 0.882 & 0.954 & 0.765 & 0.825 & 0.943 & 0.879 & 0.794 & 0.865 \\
SBI \cite{shiohara2022detecting} & 0.886 & 0.827 & 0.717 & 0.848 & 0.820 & 0.724 & 0.891 & 0.952 & 0.750 & 0.594 & 0.803 & 0.712 & 0.701 & 0.766 \\
TALL \cite{xu2023tall} & 0.831 & 0.833 & 0.693 & 0.739 & 0.774 & 0.714 & 0.699 & 0.805 & 0.651 & 0.768 & 0.863 & 0.762 & 0.616 & 0.735 \\
LSDA \cite{yan2024transcending} & 0.875 & 0.881 & 0.701 & 0.812 & 0.817 & 0.872 & 0.875 & 0.930 & 0.694 & 0.721 & 0.939 & 0.855 & 0.793 & 0.835 \\
CDFA \cite{lin2024fake} & 0.938 & 0.954 & 0.830 & 0.881 & 0.901 & 0.762 & 0.756 & 0.823 & 0.631 & 0.803 & 0.942 & 0.772 & 0.757 & 0.781 \\
\midrule 
IID$\dagger$ \cite{huang2023implicit} & 0.776 & 0.876 & 0.711 & 0.706 & 0.767 & 0.830 & 0.796 & 0.945 & 0.675 & 0.787 & 0.928 & 0.800 & 0.685 & 0.806 \\
\textbf{IID \cite{huang2023implicit} + TSRL} &\textbf{0.791} & \textbf{0.889} & \textbf{0.704} & \textbf{0.752} & \textbf{0.784} & \textbf{0.833} & \textbf{0.812} & \textbf{0.927} & \textbf{0.651} & \textbf{0.827} & \textbf{0.903} & \textbf{0.825} & \textbf{0.704} & \textbf{0.810} \\
\midrule
CLIP$\dagger$ \cite{radford2021learning} & 0.751 & 0.752 & 0.759 & 0.667 & 0.732 & 0.597 & 0.672 & 0.798 & 0.571 & 0.700 & 0.755 & 0.631 & 0.498 & 0.653 \\
\textbf{CLIP \cite{radford2021learning} + TSRL} & \textbf{0.849} & \textbf{0.732} & \textbf{0.768} & \textbf{0.724} & \textbf{0.768} & \textbf{0.598} & \textbf{0.633} & \textbf{0.827} & \textbf{0.598} & \textbf{0.728} & \textbf{0.828} & \textbf{0.663} & \textbf{0.514} & \textbf{0.674} \\
\midrule
CORE$\dagger$ \cite{ni2022core} & 0.697 & 0.868 & 0.692 & 0.759 & 0.754 & 0.855 & 0.843 & 0.936 & 0.658 & 0.741 & 0.942 & 0.834 & 0.700 & 0.814 \\
\textbf{CORE \cite{ni2022core} + TSRL} & \textbf{0.798} & \textbf{0.863} & \textbf{0.713} & \textbf{0.724} & \textbf{0.775} & \textbf{0.926} & \textbf{0.867} & \textbf{0.930} & \textbf{0.537} & \textbf{0.855} & \textbf{0.923} & \textbf{0.941} & \textbf{0.858} & \textbf{0.855} \\
\midrule
UCF$\dagger$ \cite{yan2023ucf} & 0.807 & 0.845 & 0.740 & 0.690 & 0.771 & 0.823 & 0.814 & 0.930 & 0.710 & 0.820 & 0.919 & 0.796 & 0.634 & 0.806 \\
\textbf{UCF \cite{yan2023ucf} + TSRL} & \textbf{0.843} & \textbf{0.850} & \textbf{0.738} & \textbf{0.698} & \textbf{0.782} & \textbf{0.830} & \textbf{0.821} & \textbf{0.947} & \textbf{0.718} & \textbf{0.836} & \textbf{0.910} & \textbf{0.778} & \textbf{0.652} & \textbf{0.812} \\
\midrule
ProDet$\dagger$ \cite{cheng2024can} & 0.884 & 0.878 & 0.711 & 0.801 & 0.819 & 0.869 & 0.908 & 0.959 & 0.754 & 0.699 & 0.901 & 0.809 & 0.814 & 0.839 \\
\textbf{ProDet \cite{cheng2024can} + TSRL} & \textbf{0.907} & \textbf{0.861} & \textbf{0.706} & \textbf{0.845} & \textbf{0.830} & \textbf{0.883} & \textbf{0.916} & \textbf{0.956} & \textbf{0.789} & \textbf{0.706} & \textbf{0.928} & \textbf{0.819} & \textbf{0.801} & \textbf{0.850} \\
\midrule
Effort$\dagger$ \cite{yan2024effort} & 0.871 & 0.910 & 0.863 & 0.899 & 0.886 & 0.946 & 0.839 & 0.905 & 0.987 & 0.910 & 0.967 & 0.937 & 0.870 & 0.920 \\
\textbf{Effort \cite{yan2024effort} + TSRL} & \textbf{0.901} & \textbf{0.904} & \textbf{0.882} & \textbf{0.924} & \textbf{0.903} & \textbf{0.954} & \textbf{0.902} & \textbf{0.933} & \textbf{0.983} & \textbf{0.948} & \textbf{0.975} & \textbf{0.937} & \textbf{0.904} & \textbf{0.942} \\
\bottomrule
\end{tabular*} 
\end{table*}

\subsection{Experiment Settings} 
\paragraph{Implementation Details.}For the baseline models we retrained, including IID \cite{huang2023implicit} , CLIP \cite{radford2021learning} , CORE \cite{ni2022core} , UCF \cite{yan2023ucf} , ProDet \cite{cheng2024can} , and Effort \cite{yan2024effort} , their configurations are identical to those open-sourced in DeepfakeBench \cite{DeepfakeBench_YAN_NEURIPS2023} . For the models where we employed TSRL, the original parameters from DeepfakeBench were kept consistent with the baseline, while the RL component utilized a fixed learning rate of 1e-4 for the actor and 3e-4 for the critic. Regarding evaluation metrics, we follow prior work \cite{cheng2024can, yan2024effort, lin2024fake} and adopt the video-level Area Under the Curve (AUC).

\paragraph{Datasets and Pre-processing.}Our experimental validation is conducted using the comprehensive DeepfakeBench \cite{DeepfakeBench_YAN_NEURIPS2023} framework and the recent DF40 \cite{yan2024df40} dataset. To ensure reproducibility, all dataset pre-processing strictly follows the standardized procedures outlined in these benchmarks, and we utilize the publicly available pre-processed data provided by DeepfakeBench \cite{DeepfakeBench_YAN_NEURIPS2023} .
We employ two rigorous evaluation protocols, consistent with prior work \cite{yan2024effort} : Cross-Dataset Evaluation and Cross-Method Evaluation.

\paragraph{Cross-Dataset Evaluation.}For this protocol, all models are trained exclusively on the FaceForensics++ (FF++) \cite{rossler2019faceforensics} dataset, utilizing the c23 (high-quality) video compression version as is common practice. We then evaluate the models' generalization performance on four unseen target datasets: Celeb-DF-v2 (CDF-v2) \cite{li2020celebdf} , DeepfakeDetection (DFD) \cite{dfd2020google} , Deepfake Detection Challenge (DFDC) \cite{kaggle2020dfdc} , and the DFDC preview (DFDCP) \cite{dolhansky2019dfdc} .

\paragraph{Cross-Method Evaluation.}To specifically assess robustness against novel forgery techniques within the same data domain, we evaluate the models on the DF40 dataset \cite{yan2024df40} . The data of DF40 originates from the same domain as FF++, but it is generated using a diverse set of manipulation methods not present in the FF++ training set, providing a challenging test of methodological generalization.

\paragraph{Baseline Methods.}We designate six models from DeepfakeBench \cite{DeepfakeBench_YAN_NEURIPS2023} as our core baselines: IID \cite{huang2023implicit} , CLIP \cite{radford2021learning} , CORE \cite{ni2022core} , UCF \cite{yan2023ucf} , ProDet \cite{cheng2024can} , and Effort \cite{yan2024effort} .

To ensure a direct and fair evaluation of our TSRL module, we conducted a two-stage experiment:
\begin{enumerate} [label=\textbullet, leftmargin=*, itemsep=2pt, topsep=3pt]
    \item \textbf{Baseline Reproduction:} We first retrained these six models using their official open-sourced code and the exact parameters specified in the benchmark. This process yielded our reproduced baseline results.
    \item \textbf{TSRL Integration:} We then integrated our proposed TSRL module into each of these six models and retrained them under the same conditions (with the addition of the RL-specific parameters).
\end{enumerate} 

This two-stage approach allows for a precise, ``apples-to-apples'' comparison between the performance of the original models (as reproduced by us) and their TSRL-enhanced counterparts, isolating the impact of our module.

Furthermore, we benchmark our method against a wider set of nine state-of-the-art (SOTA) models: F3Net \cite{qian2020thinking} , SPSL \cite{liu2021spatial} , SRM \cite{luo2021generalizing} , RECCE \cite{cao2022end} , SLADD \cite{chen2022self} , SBI \cite{shiohara2022detecting} , TALL \cite{xu2023tall} , LSDA \cite{yan2024transcending} and CDFA \cite{lin2024fake} . We note that we did not apply our TSRL module to this latter group, as some are not integrated into DeepfakeBench (e.g., CDFA \cite{lin2024fake} ) or utilize distinct data processing pipelines (e.g., LSDA \cite{yan2024transcending} ) that preclude straightforward integration. Their results are included for a broader contextual comparison.
\begin{table*} [t]
\centering
\caption{
    \textbf{Ablation study on the CORE model.} 
    We compare the baseline CORE with its variants: CORE + CL and our proposed CORE + TSRL. 
    All models are trained on FF++ (c23) and evaluated on the DF40 dataset. 
    We report Area Under the Curve (AUC), Accuracy (ACC), and Equal Error Rate (EER) metrics.
}
\label{tab:core_ablation}
\begin{tabular*} {\textwidth} {l l @{\extracolsep{\fill} } ccccccccc}
\toprule
Metric & Method & UniFace & BleFace & MobSwap & e4s & FaceDan & FSGAN & InSwap & SimSwap & Avg. \\
\midrule
 & CORE \cite{ni2022core} & 0.855 & 0.843 & 0.936 & 0.658 & 0.741 & 0.942 & 0.834 & 0.700 & 0.814 \\
\cmidrule{2-11}
AUC & CORE + CL & 0.862 & 0.821 & 0.913 & 0.675 & 0.774 & 0.911 & 0.851 & 0.732 & 0.817 \\
\cmidrule{2-11}
& \textbf{CORE + TSRL} & \textbf{0.926} & \textbf{0.867} & \textbf{0.930} & \textbf{0.537} & \textbf{0.855} & \textbf{0.923} & \textbf{0.941} & \textbf{0.858} & \textbf{0.855} \\
\midrule
& CORE \cite{ni2022core} & 0.725 & 0.713 & 0.825 & 0.600 & 0.648 & 0.837 & 0.718 & 0.580 & 0.706 \\
\cmidrule{2-11}
ACC & CORE + CL & 0.767 & 0.692 & 0.811 & 0.631 & 0.684 & 0.807 & 0.734 & 0.654 & 0.723 \\
\cmidrule{2-11}
& \textbf{CORE + TSRL} & \textbf{0.815} & \textbf{0.752} & \textbf{0.844} & \textbf{0.562} & \textbf{0.764} & \textbf{0.817} & \textbf{0.833} & \textbf{0.750} & \textbf{0.767} \\
\midrule
& CORE \cite{ni2022core} & 0.242 & 0.266 & 0.167 & 0.392 & 0.339 & 0.165 & 0.259 & 0.376 & 0.276 \\
\cmidrule{2-11}
EER & CORE + CL & 0.237 & 0.288 & 0.184 & 0.366 & 0.300 & 0.192 & 0.248 & 0.309 & 0.266 \\
\cmidrule{2-11}
& \textbf{CORE + TSRL} & \textbf{0.183} & \textbf{0.244} & \textbf{0.190} & \textbf{0.476} & \textbf{0.234} & \textbf{0.181} & \textbf{0.148} & \textbf{0.251} & \textbf{0.238} \\
\bottomrule
\end{tabular*}
\end{table*}
\subsection{Generalization Comparisons} 
Our analysis focuses on two comparisons: the performance delta between our reproduced baselines ($\dagger$) and their TSRL-enhanced counterparts, and the overall comparison against existing state-of-the-art (SOTA) methods.

\paragraph{Cross-Dataset Evaluation.} 
We first evaluate the models' generalization ability against unseen datasets. As shown in \textbf{Table \ref{tab:my_combined_evaluation} } , all models are trained on FF++ (c23) and tested on four unseen target datasets.
\paragraph{Impact of TSRL Module.} 
The results in \textbf{Table \ref{tab:my_combined_evaluation} } show that our TSRL module provides consistent generalization improvements across all six retrained baselines. The performance gains are particularly notable on models like \textbf{CLIP + TSRL} , which improved its average AUC from 0.732 to \textbf{0.768} (a +3.6\% gain). Our strongest model, \textbf{Effort + TSRL} , also saw its average AUC increase from 0.886 to \textbf{0.903} . This confirms that our dynamic curriculum is effective at enhancing robustness against unseen data domains.

\paragraph{Cross-method Evaluation.} 
To specifically assess the models' robustness against unseen manipulation techniques, we conduct a cross-method evaluation on the DF40 dataset \cite{yan2024df40} . As detailed in \cite{yan2024df40} , this dataset contains forgery methods not present in the FF++ training set, providing a challenging test of methodological generalization. The comprehensive results are presented in \textbf{Table \ref{tab:my_combined_evaluation} } .
\paragraph{Impact of TSRL Module.} 
Integrating our TSRL module substantially enhances the generalization capabilities of the baseline architectures. We observe consistent performance gains across diverse models. For instance, \textbf{Effort + TSRL} achieves the highest average AUC of \textbf{0.942} (a +2.2\% improvement), while \textbf{CORE + TSRL} shows the largest relative improvement (a +4.1\% gain, from 0.814 to \textbf{0.855} ). These results confirm the broad applicability and effectiveness of our module.
In summary, the cross-method evaluation confirms that TSRL provides a significant boost in robustness against novel and unseen forgery techniques, establishing a new SOTA performance level.

\begin{figure} [htbp]
\centering
\includegraphics[width=0.9\columnwidth]{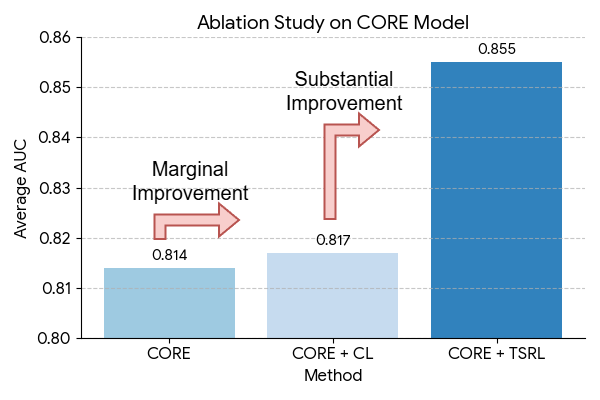} 
\caption{Visual comparison of the average AUC (on DF40) for the standard CORE baseline, CORE with a static Curriculum Learning (CL) heuristic, and the full CORE + TSRL framework. Our dynamic TSRL approach shows a significant performance gain over both other methods.} 
\label{fig:core_ablation} 
\end{figure} 
\subsection{Ablation Studies} 
We conduct a series of ablation experiments to meticulously dissect the components of our proposed framework. Our primary goal is to isolate and quantify the contributions of the heuristic curriculum (CL) and the full dynamic RL framework (TSRL). For this analysis, we use the CORE \cite{ni2022core} model as our Student backbone and the Cross-Method (DF40) evaluation protocol. The results are visually summarized in \textbf{Figure \ref{fig:core_ablation} } and detailed in \textbf{Table \ref{tab:core_ablation} } .
\begin{figure*}[t]
    \centering

    \begin{subfigure}[b]{0.49\textwidth} %
        \centering
        \includegraphics[width=\linewidth]{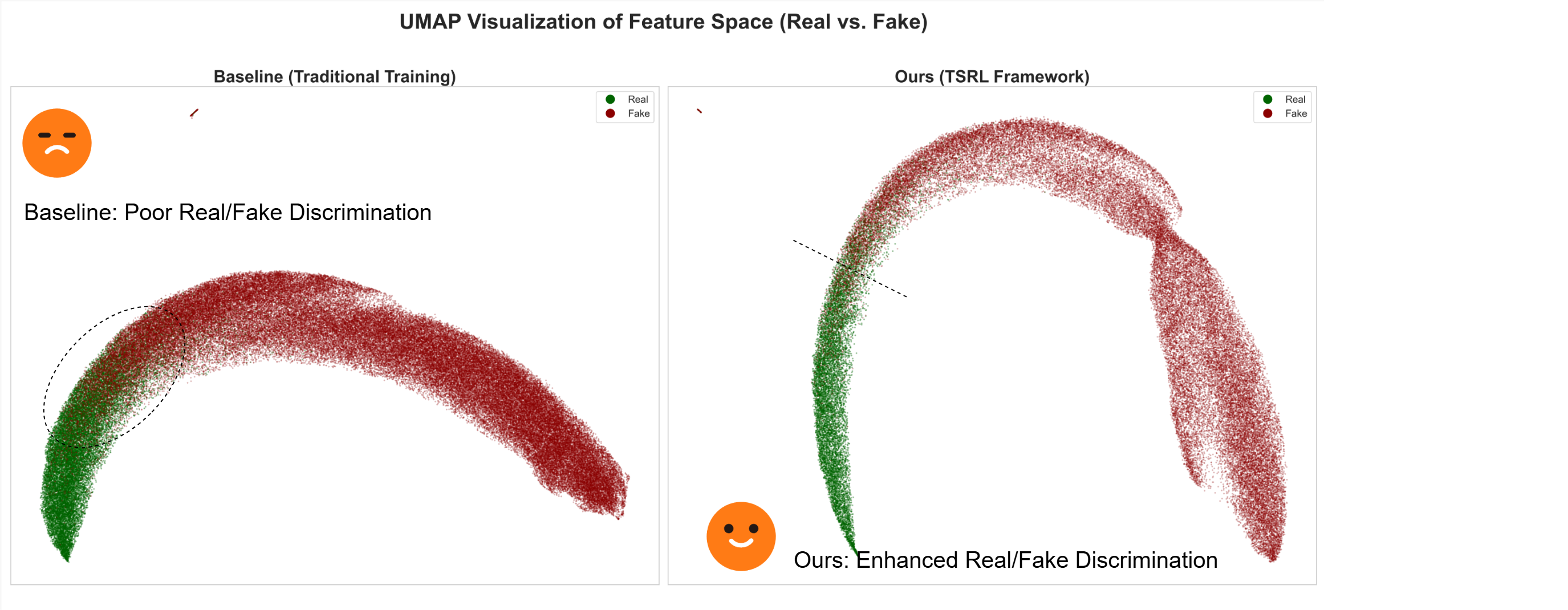}
        \caption{Fake vs Real} 
        \label{fig:fakevsreal} 
    \end{subfigure}
    \hfill 
    \begin{subfigure}[b]{0.49\textwidth} %
        \centering
        \includegraphics[width=\linewidth]{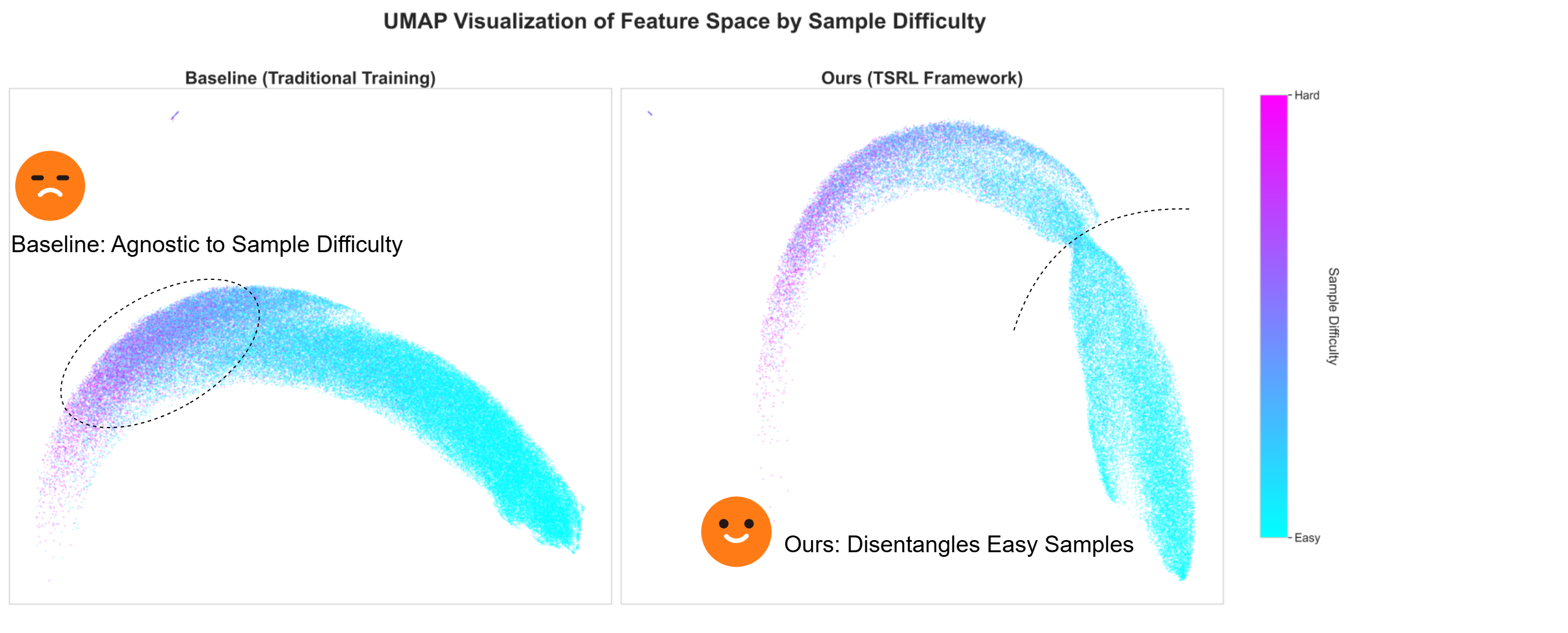}
        \caption{Fake Only} 
        \label{fig:fakeonly} 
    \end{subfigure}

    \caption{
    \textbf{UMAP visualization of feature spaces.} We present two comparative visualizations.
    \textbf{(a) Fake vs Real:} Visualization by class (Green: Real, Red: Fake). The Baseline model (left) exhibits a single manifold with heavy class overlap, indicating a confused feature space. In contrast, our TSRL framework (right) learns a perfectly disentangled representation, cleanly separating all Real samples (green arc) from all Fake samples (red arc and cloud).
    \textbf{(b) Fake Only:} Visualization of only Fake samples, colored by difficulty (Blue/Cyan: Easy, Purple/Magenta: Hard). The Baseline (left) shows a single, continuous arc with easy (blue/cyan) samples mixed throughout. Our TSRL model (right) again demonstrates superior structure, partitioning the Fake samples into two distinct clusters: a separate cloud of ``Easy Fakes'' (blue/cyan) and a primary arc of ``Hard Fakes'' (purple).
}
    \label{fig:tsne_comparison} 
\end{figure*}
\paragraph{Effects of CL.} 
We first evaluate the impact of using a static, heuristic-based curriculum, which we denote as ``CL''. This model is equivalent to our Tutor agent after the \textbf{Behavioral Cloning (BC) Initialization} step (Section \ref{sec:TSRL} , item 1) but \emph{without} any subsequent PPO optimization. The Tutor's policy is ``frozen'' to mimic the expert heuristic (e.g., assigning weights based on loss or forgetting counts), effectively serving as a sophisticated, but static, ``Self-Paced Learning'' baseline. We compare this CORE + CL model against the standard CORE baseline, which uses uniform sample weighting (i.e., all $w_i = 1.0$). As shown in \textbf{Table \ref{tab:core_ablation} } , the standard `CORE' baseline achieves an average AUC of 0.814, while the CORE + CL model achieves 0.817. This demonstrates that even a static, heuristic-driven curriculum provides a marginal performance benefit over the naive uniform-weighting paradigm. However, the small magnitude of this improvement (+0.3\%) suggests that a static policy is insufficient, as it cannot adapt to the Student's evolving learning state in real-time.

\paragraph{Effects of TSRL.} 
Next, we validate the full contribution of our proposed Tutor-Student Reinforcement Learning framework. In this setting, the Tutor agent is no longer static; it actively refines its policy $\pi$ via PPO training. The Tutor observes the rich, history-aware state $s_i$ and receives the state-change reward $r_i$ (Eq. \ref{eq:reward} ) after each weighted update, allowing it to learn a dynamic policy that maximizes the Student's immediate learning progress.

The results show a substantial performance leap. As detailed in \textbf{Table \ref{tab:core_ablation} } and \textbf{Figure \ref{fig:core_ablation} } , our full `CORE + TSRL' model achieves an average AUC of \textbf{0.855} . This is a significant \textbf{+3.8\%} improvement over the static `CORE + CL' model (0.817) and a \textbf{+4.1\%} improvement over the standard `CORE' baseline (0.814). This dramatic gain validates our core hypothesis: the dynamic, adaptive policy learned by the RL Tutor is far superior to any fixed curriculum. By optimizing its policy based on the Student's instantaneous feedback, the Tutor learns a sophisticated, adaptive curriculum that effectively prioritizes high-value, informative samples, leading to a more robust detector with significantly enhanced generalization capabilities across unseen data.

\paragraph{Qualitative Analysis of Feature Space.}
To provide a qualitative understanding of how the TSRL framework achieves its superior generalization, we visualize the feature spaces of the baseline and our full TSRL model in Figure \ref{fig:tsne_comparison}. We use UMAP to project the high-dimensional features into 2D across two comparative plots.

\textbf{Figure \ref{fig:tsne_comparison} (Fake vs Real)} visualizes all samples, colored only by their class (Green: Real, Red: Fake). The baseline model (left) exhibits a single, continuous manifold with \textbf{heavy class overlap (highlighted by the dashed circle)}, indicating a confused feature space where Real and Fake samples are poorly separated. In sharp contrast, our TSRL framework (right) learns a perfectly disentangled representation. It achieves a \textbf{clean, decisive separation (marked by the dashed line)}, organizing all Real samples into a distinct green arc, completely separate from the Fake samples.

\textbf{Figure \ref{fig:tsne_comparison} (Fake Only)} investigates the internal structure of the Fake samples, colored by their difficulty (Blue/Cyan: Easy, Purple/Magenta: Hard). The baseline (left) again shows a single, continuous structure where Easy and Hard Fakes are \textbf{intermingled (dashed circle)}. Our TSRL model (right), however, demonstrates a highly structured partition \textbf{(indicated by the dashed separation line)}. It intelligently isolates the ``Easy Fakes'' (blue/cyan) into a distinct, separate cloud, effectively ``solving'' and removing them. This leaves a primary arc composed almost entirely of ``Hard Fakes'' (purple). This visualization provides direct evidence of the RL Tutor's adaptive policy: it learns to identify and isolate trivial samples, forcing the Student to focus its representational capacity on resolving the ambiguity at the true decision boundary between Hard Reals and Hard Fakes.

%% file: sec/5_Conclusions.tex
\section{Conclusions}
\label{sec:conclusions}

This paper addresses the critical generalization gap in deepfake detection, which we attribute to static, uniform sample weighting. We reframe this process as an adaptive curriculum problem by introducing a novel Tutor-Student Reinforcement Learning (TSRL) framework. This framework models training as a Markov Decision Process, where a ``Tutor'' (a PPO agent) learns an optimal policy to dynamically assign continuous weights to samples for a ``Student'' (the deepfake detector). By applying TSRL to existing models, we establish new state-of-the-art results.

%% file: sec/6_Acknowledgments_and_Disclosure_of_Funding.tex
\section*{Acknowledgments and Disclosure of Funding}
\label{sec:Acknowledgments and Disclosure of Funding}

Our work was supported by Key Science and Technology Research Project of Xinjiang Production and Construction Corps (2025AB029), National Natural Science Foundation of China (62371350, 62501248,62372339,62472325), New Generation Artificial Intelligence-National Science and Technology Major Project (2025ZD0123603), Hubei Provincial Key R\&D Program (2025BAB021).

%% file: main.bib
@String(CVPR= {IEEE Conf. Comput. Vis. Pattern Recog.})

@String(ICCV= {Int. Conf. Comput. Vis.})

@String(ICASSP=	{ICASSP})

@String(CVPR  = {CVPR})

@String(ICCV  = {ICCV})

@article{wang2023survey,
  title={GAN-Generated Faces Detection: A Survey and New Perspectives},
  author={Wang, Xin and Guo, Hui and Hu, Shu and Chang, Ming-Ching and Lyu, Siwei},
  journal={ECAI 2023},
  pages={2533--2542},
  year={2023},
  publisher={IOS Press}
}

@article{masood2023deepfakes,
  title={Deepfakes generation and detection: state-of-the-art, open challenges, countermeasures, and way forward: Deepfakes generation and detection: state-of-the-art, open challenges, countermeasures, and way forward},
  author={Masood, Momina and Nawaz, Mariam and Malik, Khalid Mahmood and Javed, Ali and Irtaza, Aun and Malik, Hafiz},
  journal={Applied intelligence},
  volume={53},
  number={4},
  pages={3974--4026},
  year={2023},
  publisher={Springer}
}

@article{song2024quality,
  title={A quality-centric framework for generic deepfake detection},
  author={Song, Wentang and Yan, Zhiyuan and Lin, Yuzhen and Yao, Taiping and Chen, Changsheng and Chen, Shen and Zhao, Yandan and Ding, Shouhong and Li, Bin},
  journal={arXiv preprint arXiv:2411.05335},
  year={2024}
}

@inproceedings{xiao2024highquality,
  title={Are High-Quality AI-Generated Images More Difficult for Models to Detect?},
  author={Xiao, Yao and Yang, Binbin and Chen, Weiyan and Chen, Jiahao and Cao, Zijie and Dong, Ziyi and Ji, Xiangyang and Lin, Liang and Ke, Wei and Wei, Pengxu},
  booktitle={Forty-second International Conference on Machine Learning},
  year={2025}
}

@inproceedings{bengio2009curriculum,
  title={Curriculum learning},
  author={Bengio, Yoshua and Louradour, J{\'e}r{\^o}me and Collobert, Ronan and Weston, Jason},
  booktitle={Proceedings of the 26th annual international conference on machine learning},
  pages={41--48},
  year={2009}
}

@article{schulman2017ppo,
  title={Proximal policy optimization algorithms},
  author={Schulman, John and Wolski, Filip and Dhariwal, Prafulla and Radford, Alec and Klimov, Oleg},
  journal={arXiv preprint arXiv:1707.06347},
  year={2017}
}

@inproceedings{radford2021learning,
  title={Learning transferable visual models from natural language supervision},
  author={Radford, Alec and Kim, Jong Wook and Hallacy, Chris and Ramesh, Aditya and Goh, Gabriel and Agarwal, Sandhini and Sastry, Girish and Askell, Amanda and Mishkin, Pamela and Clark, Jack and others},
  booktitle={International conference on machine learning},
  pages={8748--8763},
  year={2021},
}

@article{cheng2024can,
  title={Can we leave deepfake data behind in training deepfake detector?},
  author={Cheng, Jikang and Yan, Zhiyuan and Zhang, Ying and Luo, Yuhao and Wang, Zhongyuan and Li, Chen},
  journal={Advances in Neural Information Processing Systems},
  volume={37},
  pages={21979--21998},
  year={2024}
}

@inproceedings{yan2023ucf,
  title={Ucf: Uncovering common features for generalizable deepfake detection},
  author={Yan, Zhiyuan and Zhang, Yong and Fan, Yanbo and Wu, Baoyuan},
  booktitle={Proceedings of the IEEE/CVF international conference on computer vision},
  pages={22412--22423},
  year={2023}
}

@inproceedings{yan2024transcending,
  title={Transcending forgery specificity with latent space augmentation for generalizable deepfake detection},
  author={Yan, Zhiyuan and Luo, Yuhao and Lyu, Siwei and Liu, Qingshan and Wu, Baoyuan},
  booktitle=CVPR,
  pages={8984--8994},
  year={2024}
}

@inproceedings{lin2024fake,
  title={Fake it till you make it: Curricular dynamic forgery augmentations towards general deepfake detection},
  author={Lin, Yuzhen and Song, Wentang and Li, Bin and Li, Yuezun and Ni, Jiangqun and Chen, Han and Li, Qiushi},
  booktitle={European conference on computer vision},
  pages={104--122},
  year={2024},
  organization={Springer}
}

@inproceedings{song2024towards,
  title={Towards Generic Deepfake Detection with Dynamic Curriculum},
  author={Song, Wentang and Lin, Yuzhen and Li, Bin},
  booktitle=ICASSP,
  pages={4500--4504},
  year={2024},
  organization={IEEE}
}

@inproceedings{xu2023tall,
  title={Tall: Thumbnail layout for deepfake video detection},
  author={Xu, Yuting and Liang, Jian and Jia, Gengyun and Yang, Ziming and Zhang, Yanhao and He, Ran},
  booktitle=ICCV,
  pages={22658--22668},
  year={2023}
}

@inproceedings{huang2023implicit,
  title={Implicit identity driven deepfake face swapping detection},
  author={Huang, Baojin and Wang, Zhongyuan and Yang, Jifan and Ai, Jiaxin and Zou, Qin and Wang, Qian and Ye, Dengpan},
  booktitle=CVPR,
  pages={4490--4499},
  year={2023}
}

@inproceedings{shiohara2022detecting,
  title={Detecting deepfakes with self-blended images},
  author={Shiohara, Kaede and Yamasaki, Toshihiko},
  booktitle=CVPR,
  pages={18720--18729},
  year={2022}
}

@inproceedings{cao2022end,
  title={End-to-end reconstruction-classification learning for face forgery detection},
  author={Cao, Junyi and Ma, Chao and Yao, Taiping and Chen, Shen and Ding, Shouhong and Yang, Xiaokang},
  booktitle=CVPR,
  pages={4113--4122},
  year={2022}
}

@inproceedings{chen2022self,
  title={Self-supervised learning of adversarial example: Towards good generalizations for deepfake detection},
  author={Chen, Liang and Zhang, Yong and Song, Yibing and Liu, Lingqiao and Wang, Jue},
  booktitle=CVPR,
  pages={18710--18719},
  year={2022}
}

@InProceedings{Cheng_2025_CVPR,
    author    = {Cheng, Jikang and Yan, Zhiyuan and Zhang, Ying and Hao, Li and Ai, Jiaxin and Zou, Qin and Li, Chen and Wang, Zhongyuan},
    title     = {Stacking Brick by Brick: Aligned Feature Isolation for Incremental Face Forgery Detection},
    booktitle = {Proceedings of the IEEE/CVF Conference on Computer Vision and Pattern Recognition (CVPR)},
    month     = {June},
    year      = {2025},
    pages     = {13927-13936}
}

@ARTICLE{Cheng_tip,
  author={Cheng, Jikang and Zhang, Ying and Zou, Qin and Yan, Zhiyuan and Liang, Chao and Wang, Zhongyuan and Li, Chen},
  journal={IEEE Transactions on Image Processing}, 
  title={ED4: Explicit Data-Level Debiasing for Deepfake Detection}, 
  year={2025},
  volume={34},
  number={},
  pages={4618-4630},
  doi={10.1109/TIP.2025.3588323}}

@InProceedings{ni2022core,
    author    = {Ni, Yunsheng and Meng, Depu and Yu, Changqian and Quan, Chengbin and Ren, Dongchun and Zhao, Youjian},
    title     = {CORE: COnsistent REpresentation Learning for Face Forgery Detection},
    booktitle = {Proceedings of the IEEE/CVF Conference on Computer Vision and Pattern Recognition (CVPR) Workshops},
    month     = {June},
    year      = {2022},
    pages     = {12-21}
}

@inproceedings{luo2021generalizing,
  title={Generalizing face forgery detection with high-frequency features},
  author={Luo, Yuchen and Zhang, Yong and Yan, Junchi and Liu, Wei},
  booktitle={Proceedings of the IEEE/CVF conference on computer vision and pattern recognition},
  pages={16317--16326},
  year={2021}
}

@inproceedings{liu2021spatial,
  title={Spatial-phase shallow learning: rethinking face forgery detection in frequency domain},
  author={Liu, Honggu and Li, Xiaodan and Zhou, Wenbo and Chen, Yuefeng and He, Yuan and Xue, Hui and Zhang, Weiming and Yu, Nenghai},
  booktitle={Proceedings of the IEEE/CVF conference on computer vision and pattern recognition},
  pages={772--781},
  year={2021}
}

@inproceedings{afchar2018mesonet,
  title={Mesonet: a compact facial video forgery detection network},
  author={Afchar, Darius and Nozick, Vincent and Yamagishi, Junichi and Echizen, Isao},
  booktitle={2018 IEEE international workshop on information forensics and security (WIFS)},
  pages={1--7},
  year={2018},
  organization={IEEE}
}

@article{goodfellow2014generative,
  title={Generative adversarial nets},
  author={Goodfellow, Ian J and Pouget-Abadie, Jean and Mirza, Mehdi and Xu, Bing and Warde-Farley, David and Ozair, Sherjil and Courville, Aaron and Bengio, Yoshua},
  journal={Advances in neural information processing systems},
  volume={27},
  year={2014}
}

@inproceedings{nadimpalli2022improving,
  title={On improving cross-dataset generalization of deepfake detectors},
  author={Nadimpalli, Aakash Varma and Rattani, Ajita},
  booktitle={Proceedings of the IEEE/CVF conference on computer vision and pattern recognition},
  pages={91--99},
  year={2022}
}

@inproceedings{li2025raidx,
  title={RAIDX: A Retrieval-Augmented Generation and GRPO Reinforcement Learning Framework for Explainable Deepfake Detection},
  author={Li, Tianxiao and Huang, Zhenglin and Wen, Haiquan and He, Yiwei and Lyu, Shuchang and Wu, Baoyuan and Cheng, Guangliang},
  booktitle={Proceedings of the 33rd ACM International Conference on Multimedia},
  pages={11746--11755},
  year={2025}
}

@inproceedings{wang2025dfbench,
  title={Dfbench: Benchmarking deepfake image detection capability of large multimodal models},
  author={Wang, Jiarui and Duan, Huiyu and Wang, Juntong and Jia, Ziheng and Yang, Woo Yi and Zhu, Xiaorong and Zhao, Yu and Qian, Jiaying and Xing, Yuke and Zhai, Guangtao and others},
  booktitle={Proceedings of the 33rd ACM International Conference on Multimedia},
  pages={12666--12673},
  year={2025}
}

@inproceedings{yan2024effort,
  title={Orthogonal Subspace Decomposition for Generalizable AI-Generated Image Detection},
  author={Yan, Zhiyuan and Wang, Jiangming and Jin, Peng and Zhang, Ke-Yue and Liu, Chengchun and Chen, Shen and Yao, Taiping and Ding, Shouhong and Wu, Baoyuan and Yuan, Li},
  booktitle={International Conference on Machine Learning},
  pages={70268--70288},
  year={2025},
  organization={PMLR}
}

@inproceedings{li2021frequency,
  title={Frequency-aware discriminative feature learning supervised by single-center loss for face forgery detection},
  author={Li, Jiaming and Xie, Hongtao and Li, Jiahong and Wang, Zhongyuan and Zhang, Yongdong},
  booktitle={Proceedings of the IEEE/CVF conference on computer vision and pattern recognition},
  pages={6458--6467},
  year={2021}
}

@inproceedings{zhu2021face,
  title={Face forgery detection by 3d decomposition},
  author={Zhu, Xiangyu and Wang, Hao and Fei, Hongyan and Lei, Zhen and Li, Stan Z},
  booktitle={Proceedings of the IEEE/CVF conference on computer vision and pattern recognition},
  pages={2929--2939},
  year={2021}
}

@inproceedings{li2020facexray,
  title={Face x-ray for more general face forgery detection},
  author={Li, Lingzhi and Bao, Jianmin and Zhang, Ting and Yang, Hao and Chen, Dong and Wen, Fang and Guo, Baining},
  booktitle={Proceedings of the IEEE/CVF conference on computer vision and pattern recognition},
  pages={5001--5010},
  year={2020}
}

@inproceedings{zhao2021learning,
  title={Learning self-consistency for deepfake detection},
  author={Zhao, Tianchen and Xu, Xiang and Xu, Mingze and Ding, Hui and Xiong, Yuanjun and Xia, Wei},
  booktitle={Proceedings of the IEEE/CVF international conference on computer vision},
  pages={15023--15033},
  year={2021}
}

@inproceedings{qian2020thinking,
  title={Thinking in frequency: Face forgery detection by mining frequency-aware clues},
  author={Qian, Yuyang and Yin, Guojun and Sheng, Lu and Chen, Zixuan and Shao, Jing},
  booktitle={European conference on computer vision},
  pages={86--103},
  year={2020},
  organization={Springer}
}

@article{yan2024df40,
  title={Df40: Toward next-generation deepfake detection},
  author={Yan, Zhiyuan and Yao, Taiping and Chen, Shen and Zhao, Yandan and Fu, Xinghe and Zhu, Junwei and Luo, Donghao and Wang, Chengjie and Ding, Shouhong and Wu, Yunsheng and others},
  journal={Advances in Neural Information Processing Systems},
  volume={37},
  pages={29387--29434},
  year={2024}
}

@inproceedings{DeepfakeBench_YAN_NEURIPS2023,
 author = {Yan, Zhiyuan and Zhang, Yong and Yuan, Xinhang and Lyu, Siwei and Wu, Baoyuan},
 booktitle = {Advances in Neural Information Processing Systems},
 editor = {A. Oh and T. Neumann and A. Globerson and K. Saenko and M. Hardt and S. Levine},
 pages = {4534--4565},
 title = {DeepfakeBench: A Comprehensive Benchmark of Deepfake Detection},
 url = {https://proceedings.neurips.cc/paper_files/paper/2023/file/0e735e4b4f07de483cbe250130992726-Paper-Datasets_and_Benchmarks.pdf},
 volume = {36},
 year = {2023}
}

@inproceedings{rossler2019faceforensics,
  title={Faceforensics++: Learning to detect manipulated facial images},
  author={Rossler, Andreas and Cozzolino, Davide and Verdoliva, Luisa and Riess, Christian and Thies, Justus and Nie{\ss}ner, Matthias},
  booktitle={Proceedings of the IEEE/CVF international conference on computer vision},
  pages={1--11},
  year={2019}
}

@inproceedings{li2020celebdf,
  title={Celeb-df: A large-scale challenging dataset for deepfake forensics},
  author={Li, Yuezun and Yang, Xin and Sun, Pu and Qi, Honggang and Lyu, Siwei},
  booktitle={Proceedings of the IEEE/CVF conference on computer vision and pattern recognition},
  pages={3207--3216},
  year={2020}
}

@misc{kaggle2020dfdc,
  title={{Deepfake detection challenge}},
  author={Kaggle},
  year={2020},
  howpublished = {\url{https://www.kaggle.com/c/deepfake-detection-challenge}},
  note = {Accessed: 2021-04-24}
}

@misc{dfd2020google,
  title={{DFD}},
  author={Google AI Blog},
  year={2020},
  howpublished = {\url{https://ai.googleblog.com/2019/09/contributing-data-to-deepfake-detection.html}},
  note = {Accessed: 2021-04-24}
}

@article{dolhansky2019dfdc,
  title={The deepfake detection challenge (dfdc) preview dataset},
  author={Dolhansky, Brian and Howes, Russ and Pflaum, Ben and Baram, Nicole and Ferrer, Cristian Canton},
  journal={arXiv preprint arXiv:1910.08854},
  year={2019}
}
